\documentclass[conference]{IEEEtran}
\usepackage{graphicx}
\usepackage{multicol,caption}
\usepackage{geometry}
\usepackage{datetime}
\usepackage[natbib=true, style=numeric,sorting=none]{biblatex}
\usepackage{subcaption}
\usepackage{float}
\usepackage{multirow}
\usepackage{notoccite}
\usepackage{amsmath,amssymb,amsfonts}
\usepackage{algorithmic}
\usepackage{textcomp}
\usepackage{xcolor}
\def\BibTeX{{\rm B\kern-.05em{\sc i\kern-.025em b}\kern-.08em
    T\kern-.1667em\lower.7ex\hbox{E}\kern-.125emX}
    }
    
\title{Board-to-Board: Evaluating Moonboard Grade Prediction Generalization \\\vspace*{20pt} \normalsize  \today{}}

\author{
\IEEEauthorblockN{Daniel Petashvili}
\IEEEauthorblockA{\textit{School of Computer and Mathematical Sciences} \\
\textit{University of Adelaide}\\
a1800310@adelaide.edu.au}
\and
\IEEEauthorblockN{Matthew Rodda}
\IEEEauthorblockA{\textit{School of Computer and Mathematical Sciences} \\
\textit{University of Adelaide}\\
a1773620@adelaide.edu.au}
}

\date{November 2023}

\begin{document}

\maketitle

\section{Abstract}
Bouldering is a sport where athletes aim to climb up an obstacle using a set of defined holds called a route. Typically routes are assigned a grade to inform climbers of its difficulty and allow them to more easily track their progression. However, the variation in individual climbers technical and physical attributes and many nuances of an individual route make grading a difficult and often biased task. In this work, we apply classical and deep-learning modelling techniques to the 2016, 2017 and 2019 Moonboard datasets, achieving performance on par with state of the art methods at 0.86 MAE and 1.12 RMSE. We achieve this performance on a feature-set that does not require decomposing routes into individual moves, which is common in literature requiring manual effort and introducing bias. We also demonstrate the generalization capability of this model between editions and introduce a novel vision-based method of grade prediction. While the generalization performance of these techniques is below human level performance currently, we propose these methods as a basis for future work. Such a tool could be implemented in pre-existing mobile applications and would allow climbers to better track their progress and assess new routes with reduced bias.

\section{Introduction}
Bouldering is a sport which has exploded in popularity over recent years. This can be attributed to the influence of social media platforms and the inclusion of bouldering as an event in the Olympic games. This sport requires climbing a wall with a set of predefined holds called a route.

\begin{figure*}[t!]
    \centering
    \begin{subfigure}[t]{0.25\linewidth}
        \centering
        \includegraphics[width=\linewidth]{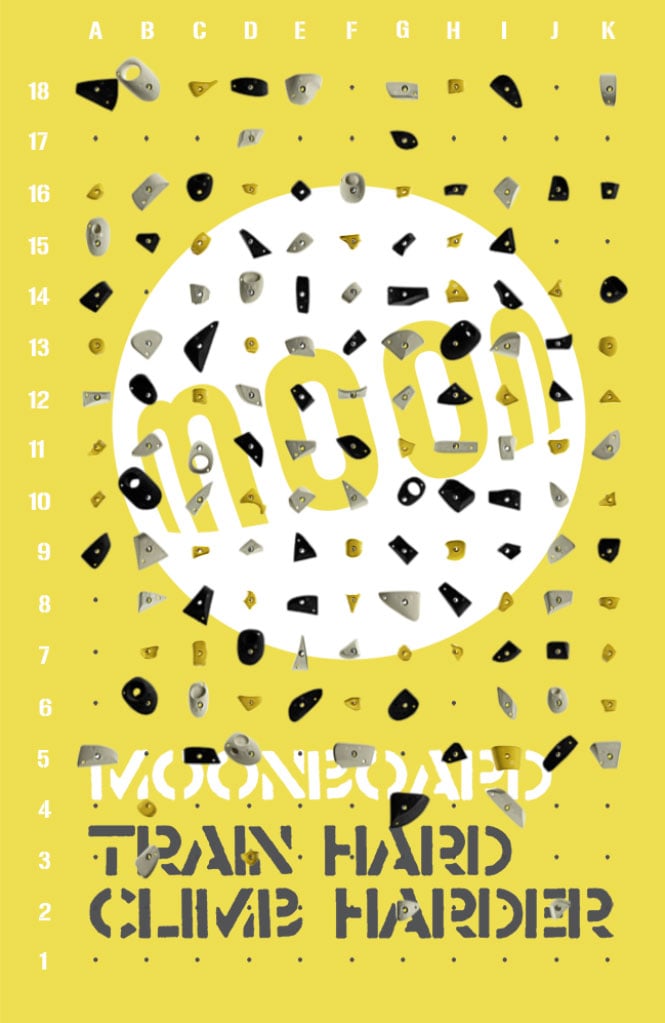}
        \caption{Moonboard 2016 Edition}
        \label{subfig:Moonboard2016}
    \end{subfigure}%
    ~ 
    \begin{subfigure}[t]{0.25\linewidth}
        \centering
        \includegraphics[width=\linewidth]{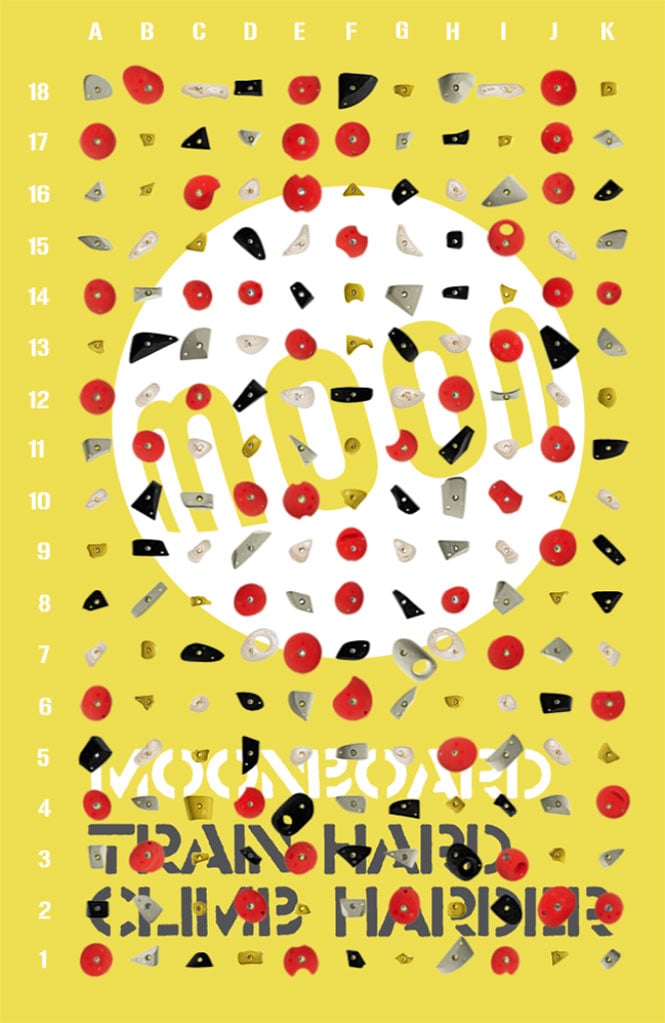}
        \caption{Moonboard 2017 Edition}
        \label{subfig:Moonboard2017}
    \end{subfigure}
    ~ 
    \begin{subfigure}[t]{0.25\linewidth}
        \centering
        \includegraphics[width=\linewidth]{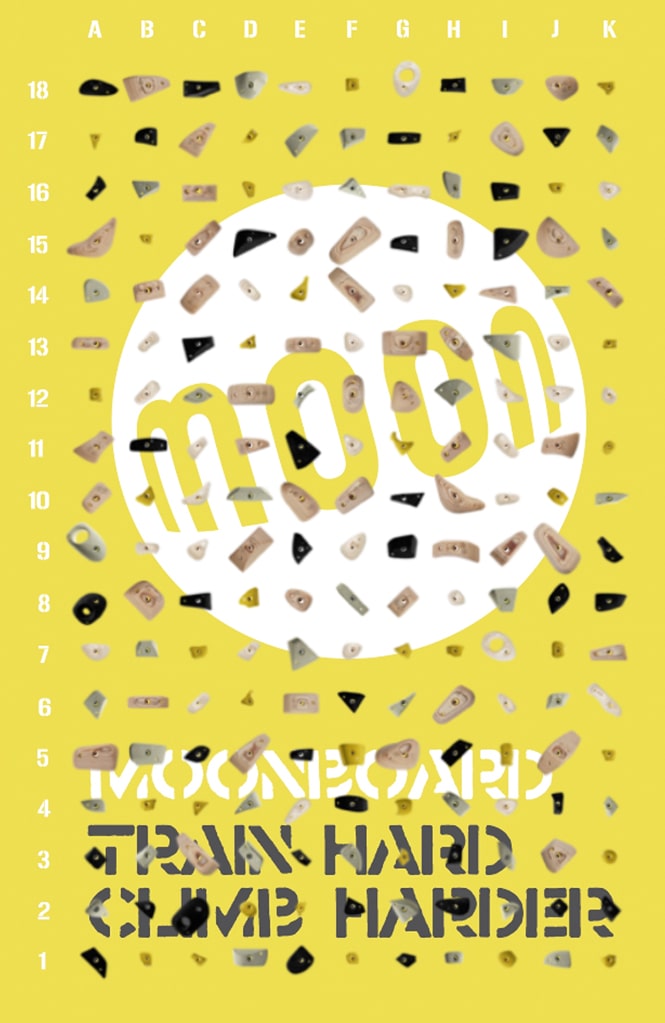}
        \caption{Moonboard 2019 Edition}
        \label{subfig:Moonboard2019}
    \end{subfigure}
    \captionof{figure}{Moonboard Editions with Year}
    \label{fig:configs}
\end{figure*}

The Moonboard, shown in Figure \ref{fig:configs}, is a standardised training wall for bouldering and is used by climbers internationally. Using a smartphone application, climbers can create their own routes or try other climber's routes. The displayed difficulty of each route is based on a majority voting system.

Standardised difficulty grading systems for bouldering routes exist and are used globally, with the V-scale and Font-scale being the most widely accepted. Due to its higher granularity, in this work we focus on the Font-scale. Many factors about a route such as the size, shape and number of holds may be indicative of its difficulty. Subtle changes in a route can have a significant impact on the difficulty. This is complicated further by an individual climber's technical strengths and physical attributes; while one climber may comfortably find a hold within reach, another shorter climber may need to jump and catch a hold. These variations can lead to large differences in the perceived difficulty of a route.

Little objective criteria exists to grade a climbing route, so climbers must rely on their subjective experience. This makes it difficult to grade routes in a fair and consistent manner. Variation in boulder grades is most apparent when analysing disjoint communities of climbers that tend to reach a different internal consensus about what is required of a specific grade. We can observe a real example of this effect in Japan, which is notorious for underrating the difficulty of routes.

To improve the objectivity of grading, we propose a machine learning algorithm which predicts the difficulty of a route. We consider better than human level performance to be a practical goal of such a model, which could then be developed into a tool to suggest an accurate grade for new boulders to a human user. Such a tool could improve the accessibility of bouldering, as grades could become more consistent across different communities. This is particularly important for novice climbers track their progression, as a poorly graded routes can easily discourage them from continuing to participate.

To achieve the goal of this project we explored a variety of machine learning approaches for boulder grade prediction. This exploration included classical and deep learning approaches. Our modelling and analysis was performed on the Moonboard due to its standardisation and use in the relevant literature. Through this, we evaluated each technique and identified its applicability to this problem. This report discusses how we progressed towards the project goal, the novelty and significance of our work and our learning.

\section{Related Work}
Dobles \cite{Dobles2017MachineLM} performed experimentation on the Moonboard with naive bayes, softmax regression and CNN classification. The CNN showed the greatest performance with an MAE of 1.40. This was performed using a one-hot encoding of the Moonboard.

Kempen \cite{kempen2018fair} used human understanding of the routes to generate a more detailed dataset. Rather than providing which holds are available, each route is converted to a sequence of moves which a climber would need to perform. This mimics how humans approach bouldering, where they think about different sequences and what would be the easiest way to complete the route. Performing this process manually is time consuming, which can limit the size of the data set. Such a method is also reliant on the expertise of the labeller as they must solely decide the difficulty of a range of different moves. To overcome these issues, a simplifying assumption was used, splitting the grades into a binary classification of easy or hard. They achieved a classification accuracy of approximately 60\% using k-fold cross validation, narrowly outperforming a randomly guessing binary classifier.

Duh\cite{Duh2020RecurrentNN} created a grade prediction pipeline which programmatically converted the allowed holds to the most probable sequence of moves. Automatically generating the move sequence is advantageous as it allows for large datasets to be used. Furthermore, the generation process was carefully evaluated to ensure that it produces the correct sequence of moves. These two factors increase the robustness of the model when compared to Kempen's human driven approach. The sequence of moves is by a beam search algorithm, common in sequence prediction tasks. The beam search algorithm uses a success score which is calculated based on the distance between holds and the relative difficulty of holds for a given move of the route.

Using the move sequence as input, they created an recurrent neural network (RNN) which predicted grade. The RNN achieved 46.7\% on exact matches and 84.7\% on matches within 1 grade; compared to human level performance of 45\% and 85\%. They also found that the move sequence conversion was a significant part of their success; as a RNN with the holds as input only reached 34.7\% exact matches. These results demonstrate the advantages of deep learning approaches in this context. They also indicate the power of using a hold to move conversion step.

Stapel \cite{stapel2023} similarly creates a grade prediction pipeline that decomposes full routes into individual moves. Each route is converted to an ordered sequence of moves, which are then individually assessed and then aggregated into an overall grade prediction. The move sequence is chosen by determining the easiest sequence according to a heuristic created by Stapel that considers the size of the included holds, the distance between holds, presence of footholds and intuitiveness of the move. Each move in that sequence is then evaluated by a suite of binary classifiers, each predicting whether a move is more or less difficult than a chosen V-scale threshold. Each of these probability predictions is then used by a final classifier as a feature to predict the overall grade.

Using the 2016 Moonboard, Stapel achieves 46.5\% exact matches and 80.1\% within one grade on a 25\% held out test set from a total dataset of 30641 routes. This best performance was achieved with a histogram based gradient boosting classifier and is on par with \cite{Duh2020RecurrentNN}. However, a limitation of this method is the requirement of a dataset of moves for implementing the ordered sequence prediction step. This dataset required the author manually label 250 routes with their expected move sequence and would need to be performed again to apply the method to a new board.

\section{Novelty and Significance}
The established literature focuses on techniques which involve preprocessing holds into a sequence of moves, and when done effectively, grade prediction performance is shown to be improved. However, these methods introduce the assumptions that there is only one easiest sequence for all climbers and that the manual labeller has correctly identified it. In reality, different climbers often find novel sequences that suit their strengths and these sequences are not easily determined. To address these concerns, we perform grade prediction on routes without preprocessing individual moves and demonstrate its effectiveness.

The relevant literature so far demonstrates grade prediction on a board configuration given a training set on the same configuration. This is a highly limiting assumption as a climber must build a corpus of graded routes on their board configuration before expecting accurate predictions. The quality of such grade predictions is also entirely dependant on the accuracy of this new corpus. As such, in this work we explore the generalization of a model trained on a given Moonboard edition to a new edition. This evaluation reveals to what extent a machine learning model can learn spatial relationships to predict grades rather than information about specific holds.

To the best of our knowledge, there are also no contributions in this area using pure computer vision approaches. Such an approach could generalize to boulder grade prediction outside of narrow constraints of a standardized training board. This could offer a framework to solve boulder grade prediction problems with novel holds and complex 3D wall geometries that the Moonboard dataset does not explore. In this work, we provide preliminary results from training computer vision backbones to predict boulder grades from images.

\section{Evaluation}
For this boulder grade prediction task we propose evaluation using Mean Absolute Error (MAE) and Root Mean Squared Error (RMSE). These metrics are extremely common amongst regression tasks in machine learning literature in general and have been used directly in \cite{Dobles2017MachineLM}. Use of these common metrics allows comparison of our results with some relevant literature. MAE measures the error between predicted and true grades uniformly, whereas the RMSE increases more quickly with errors larger than 1. The combination of these two metrics is useful as comparison between MAE and RMSE can indicate tendencies within the model. When RMSE is greater than MAE we can conclude that our model is prone to making a few highly incorrect predictions. Conversely, when RMSE is less than MAE, we know the model is rarely perfectly correct and is likely making predictions with some uniform error less than one grade. The mathematical definition of these two evaluation metrics are:

\begin{align}
\label{eq:MAE}
MAE = \sum_{i=1}^{D}|x_i-y_i|
\end{align}

\begin{align}
\label{eq:RMSE}
RMSE = \sqrt{\sum_{i=1}^{D}\frac{(x_i-y_i)^2}{D}}
\end{align}

\section{Datasets}
We train and evaluate our modelling techniques against 3 datasets all collected from the popular Moonboard training board. Each dataset is collected from a different edition of the Moonboard, with the number of routes shown in Table \ref{tab:datasets}. Between each of the 3 Moonboard editions evaluated, 142 holds are shared while an additional 56 holds are added that vary between the most recent 2017 and 2019 editions, as shown in Figure \ref{fig:configs}.

 \begin{center}
     \begin{tabular}{|c|c|c|c|}
     \hline
        Edition (Year) & Holds & Routes & Benchmarks \\ \hline
        2016 & 142 & 28568 & 545 \\ \hline
        2017 & 198 & 24083 & 336 \\ \hline
        2019 & 198 & 11375 & 383 \\ \hline
    \end{tabular}
    \captionof{table}{Number of Holds, Routes and Benchmarks per Edition}
    \label{tab:datasets}
\end{center}

These datasets were scraped from the Moonboard API, where a user of the board can publish and grade their own routes publicly for others to try. While a single user grading these routes makes the label unreliable, other users who complete the route can suggest their perceived grade to update the difficulty. When building our dataset, we select only routes with 5 or more ascents such that the grade has reached some consensus.

Each dataset contains a relatively small number of 'Benchmark' routes. These are exemplar routes that experience considerably more ascents, and consequentially ratings. The grades for benchmark routes are highly agreed upon and verified by experts who maintain the Moonboard tool. For this reason, we select these routes as testing sets for our evaluation. The remaining non-benchmark routes comprise the training set.

Each element of these datasets is a list of hold coordinates, supervised by the consensus grade of all of its ascents. From this list of hold coordinates we construct a 18x11 dimensional one-hot encoded feature vector to represent the route. An example of this conversion is shown in Figure \ref{fig:one-hot}, showing an image of the route from the Moonboard website, the associated list of holds, and our one-hot feature vector representing some spatial understanding of the included holds.

\begin{center}
    \centering
    \includegraphics[width=\linewidth]{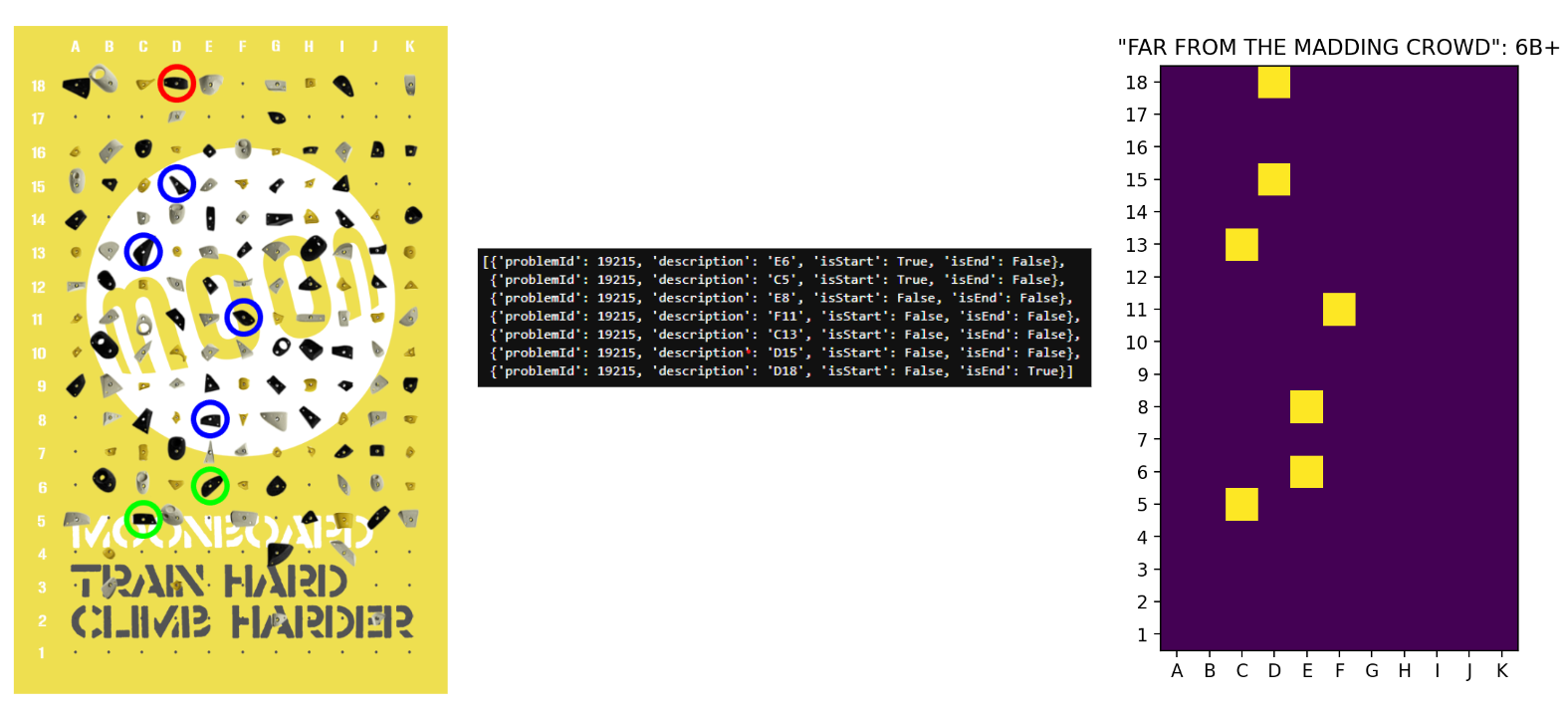}
    \captionof{figure}{One-Hot Feature Vector from List of Holds}
    \label{fig:one-hot}
\end{center}

As all three Moonboard editions share the same dimension, we can apply the same modelling algorithms across all three datasets without change. However, holds changing position across the datasets make generalization of performance across datasets challenging. For example, modelling one dataset, an algorithm may learn that a hold in a specific position is associated with high difficulty routes. Testing on a new board configuration, that same position may be replaced with a relatively easy hold leading to incorrect grade predictions.

Each dataset is highly imbalanced in its representation of each target difficulty. There are significantly more low grade routes than higher grade ones. Such an imbalance is expected as there are many more novice and intermediate climbers than experts. This is most apparent in the non-benchmark training sets, however this is still true to a lesser extent in the testing sets, as shown if Figure \ref{fig:distributions}. The 2016 edition also limits routes to a minimum grade of 6B+, whereas 2017 and 2019 editions allow labelling routes to a more inclusive minimum grade of 6A+.

\begin{center}
    \centering
    \includegraphics[width=\linewidth]{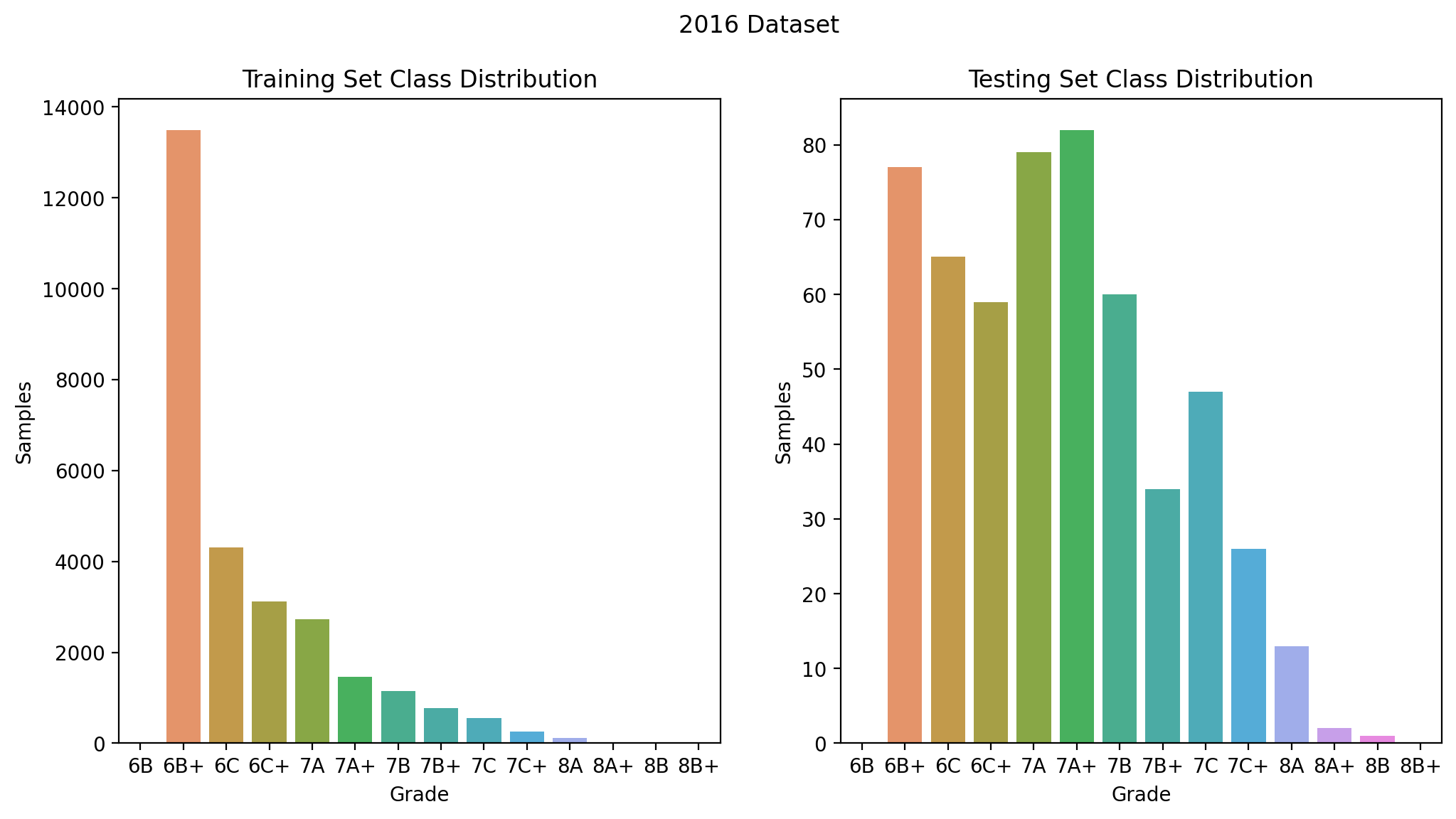}
    \includegraphics[width=\linewidth]{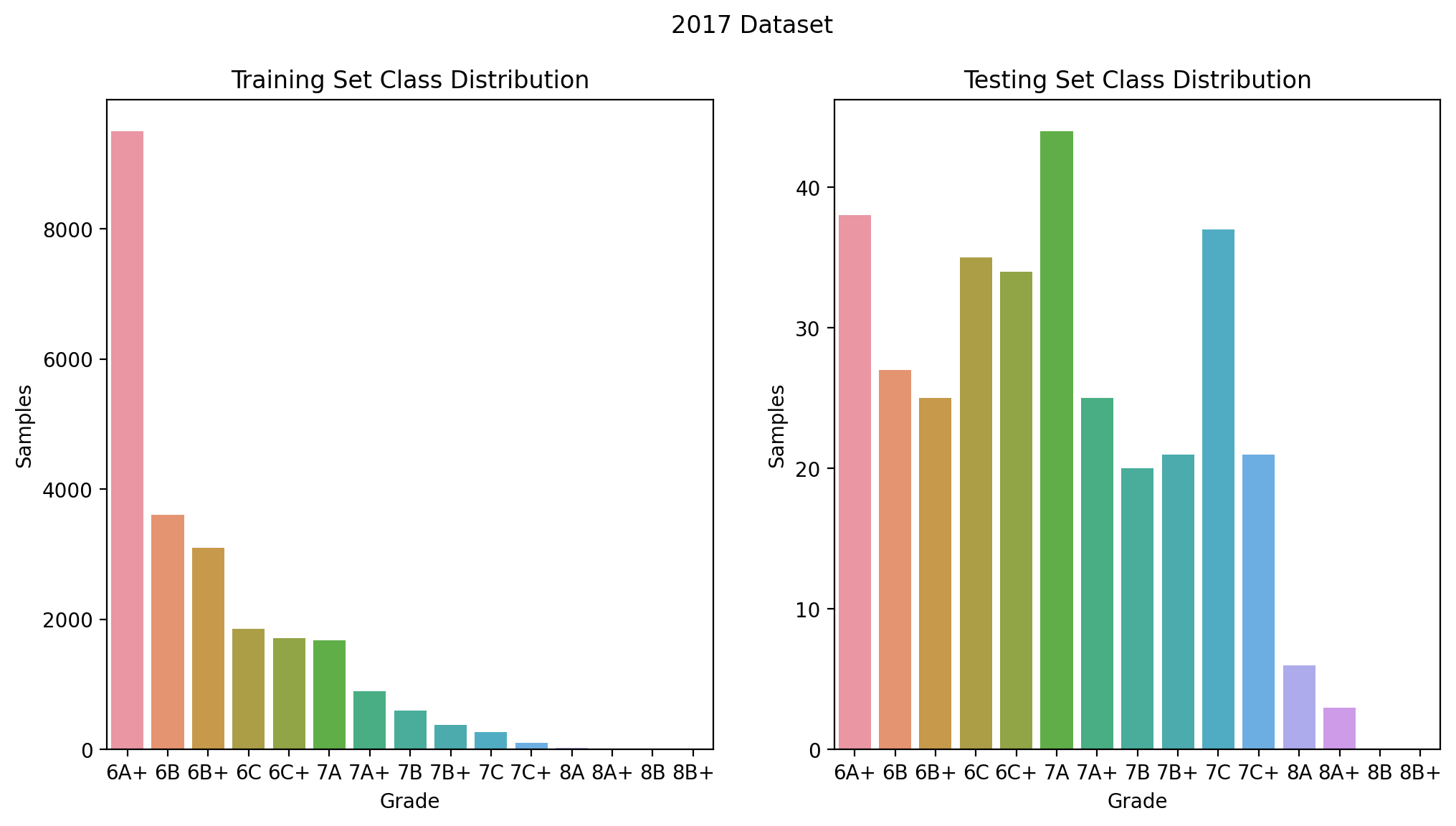}
    \includegraphics[width=\linewidth]{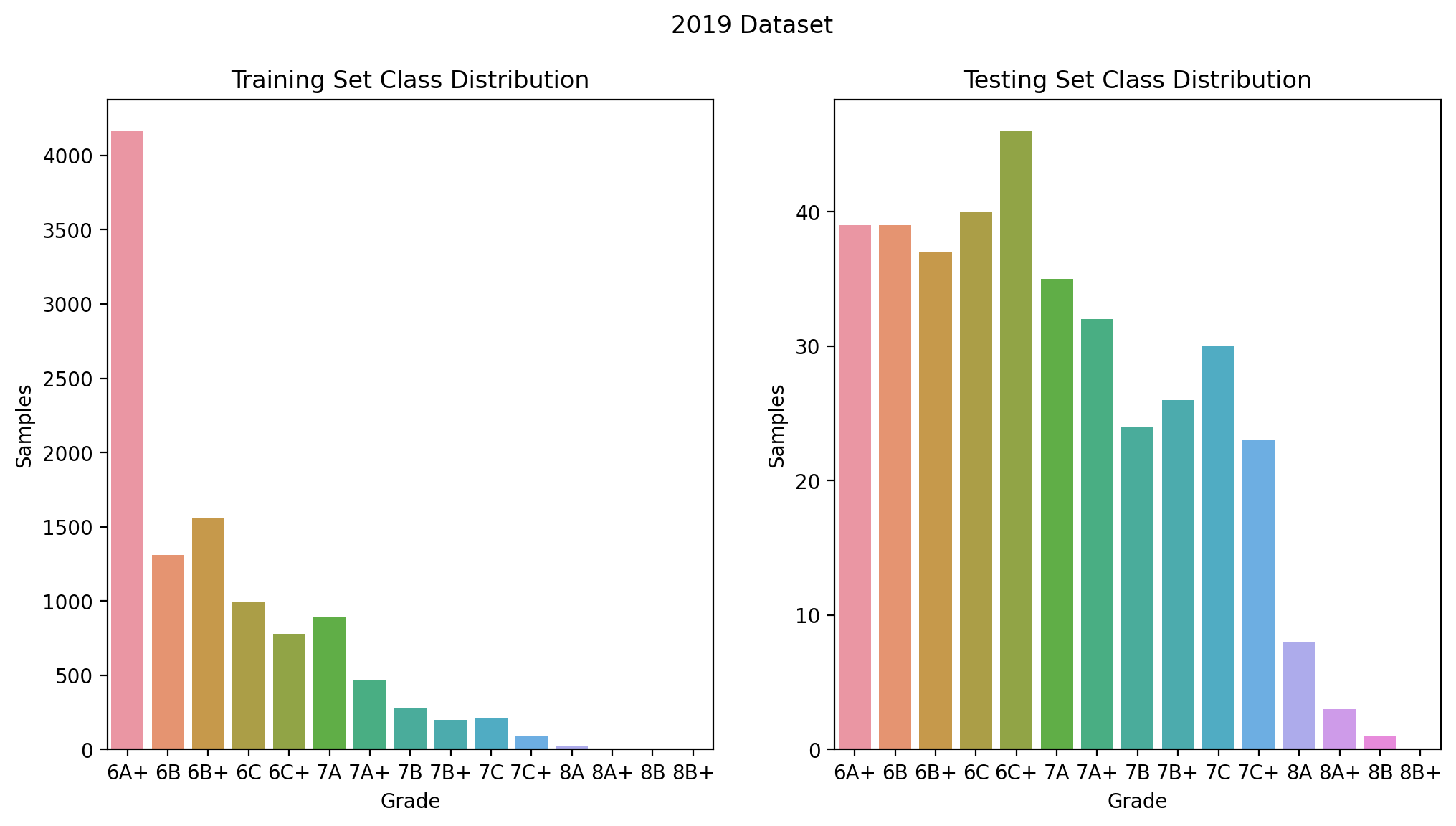}
    \captionof{figure}{Target Distribution between training and testing sets per dataset}
    \label{fig:distributions}
\end{center}

\section{Methods}
In this section we explore the performance of different modelling techniques, hyperparameter tuning and feature sets constructed from our dataset. We start with classical machine learning to build baseline performance for our modelling. We then explore deep-learning algorithms to improve our performance on the same feature set. Finally, we explore computer vision models trained on images of routes for grade prediction.

\subsection{Classical Machine Learning}
We start by evaluating relatively simple classical models on our grade prediction task. The fast training and evaluation of such classical models allowed us to quickly gain insight and establish a baseline performance to improve upon. We train and evaluate a suite of 6 classical regression models. First, we evaluate three linear modelling techniques: the linear regression (LR), linear regression with L2 normalization (LR with L2) and support vector regression (SVR). Next, we evaluate three non-linear regression models: radial basis function support vector regression (RBF SVR), decision tree regressor (DTR) and extreme gradient boosted decision tree regressor (XGBR).

We train and evaluate this suite of 6 models against the Moonboard 2016 dataset, with performance shown in Figure \ref{fig:classical}. The three linear models achieve similar performance on the training set with 1.03 and 1.40 MAE and RMSE respectively. Such low performance on the training set indicates that the assumption of linear contribution of our features to the predicted grade is insufficient. While linear modelling achieves poor performance, the LR with L2 is particularly useful for its ease of interpretation. As shown in Figure \ref{fig:weights}, the weights of this model give some intuition as to the difficulty of the routes it is included in. Inspection of the highest weighted positions and referencing Figure \ref{fig:configs}(a), reveals that smaller holds often increase the predicted difficulty.

\begin{center}
    \centering
    \includegraphics[width=0.8\linewidth]{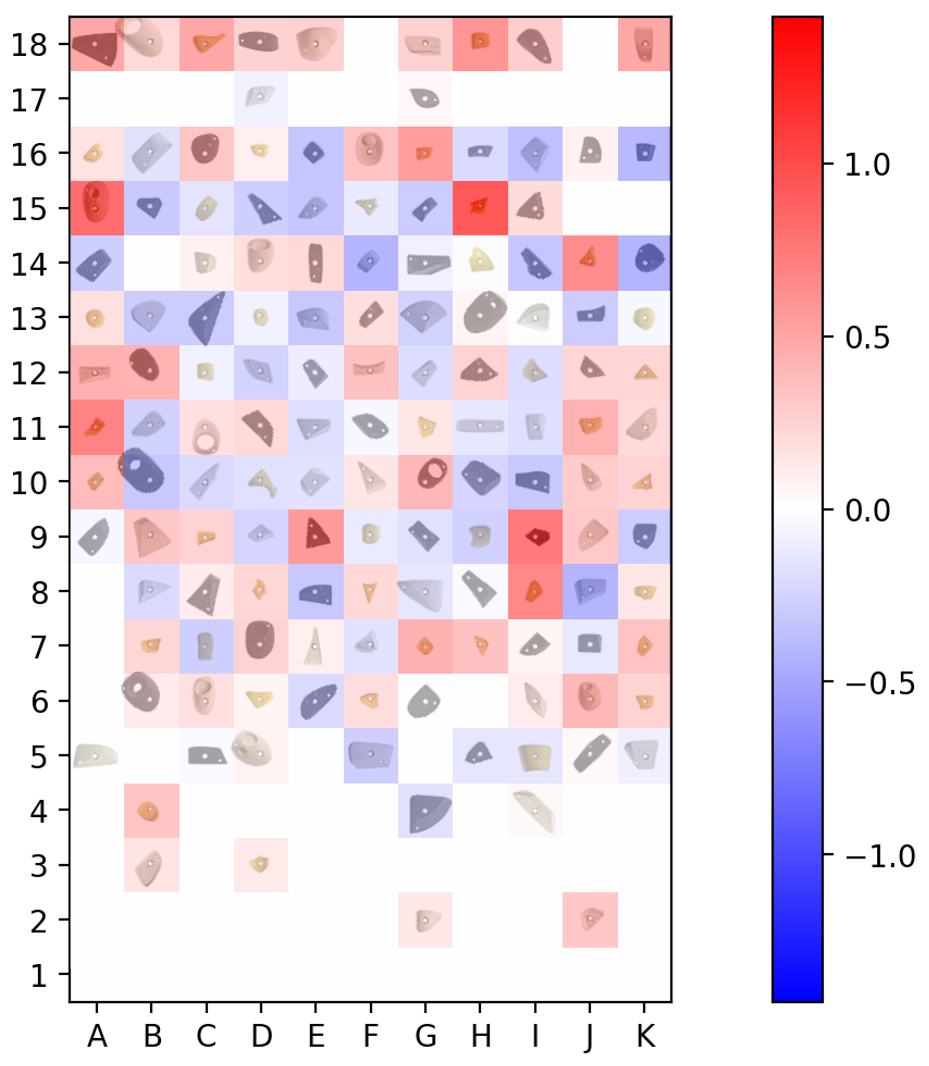}
    \captionof{figure}{Weight per Hold on 2016 Dataset from LR w/ L2 normalization. Red indicates a hold associated with higher difficulty, white indicates a hold of average difficulty and blue indicating an easier than average hold.}
    \label{fig:weights}
\end{center}

Individual holds are rarely used in isolation, instead an athlete must move between holds. As such, the predicted grade should not be a linear weighting of which holds are included, instead it must be made with some understanding of the complex interaction of different holds and how they may be used together.

These interactions perhaps explain the lower errors achieved by non-linear modelling. Without regularization, the DTR overfit to the training set, and failed to generalize to the testing set. This overfitting is expected, as an unconstrained decision tree can perfectly fit any training set. While the XGBR algorithm is also a tree building model, it utilizes boosting which iteratively adds decision trees, with each addition correcting the errors from the last tree. This boosting is a form of regularization that leads to worse performance on the training set, but much better generalization to the testing set. Best performance on the test set is achieved by the RBF SVR. Using the radial basis function kernel is equivalent to transforming your feature space into an infinite dimensional feature space, where each component of the feature vector is a product of powers of the components in the original feature space. With one-hot encoded features, these products represent combinations of all possible subsets of holds. For example, the first degree product considers only the presence of a single hold. Next, the second degree polynomial considers the squared contribution of features (which for one-hot encoded features is trivial, remaining either zero or one) and the product of all combinations of two features. Again, for one-hot encoded features, this product is zero when only one or neither of the holds is present, and one when both are present. Such a product is similar to performing a logical AND between the considered holds. This continues to infinite degree products, where all possible combinations of included holds have been considered.

This could be one explanation for why the RBF SVM achieves best performance of 0.98 MAE on the testing set with 1.28 RMSE, meeting our target human level performance.

\begin{center}
    \centering
    \includegraphics[width=\linewidth]{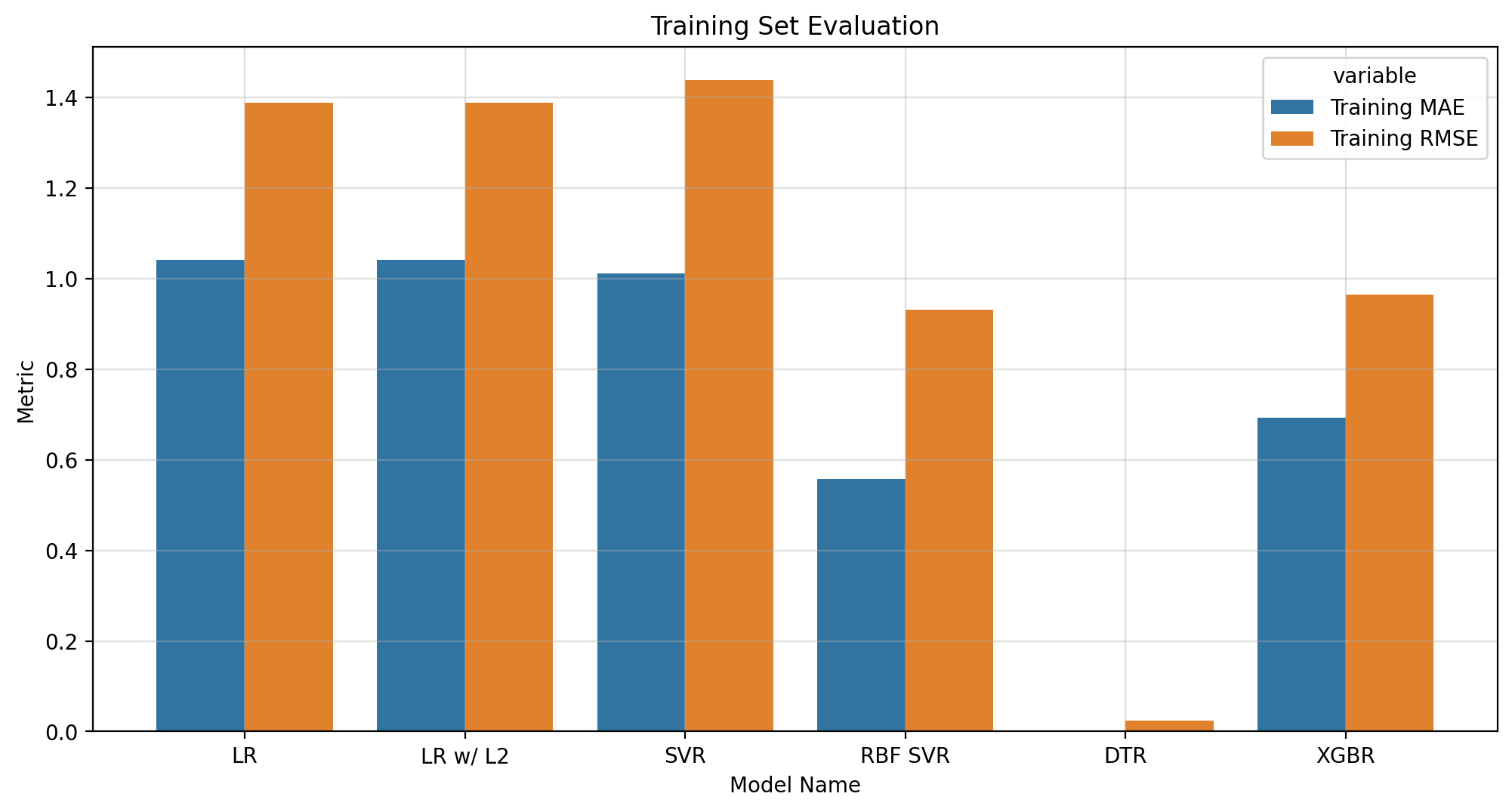}
    \includegraphics[width=\linewidth]{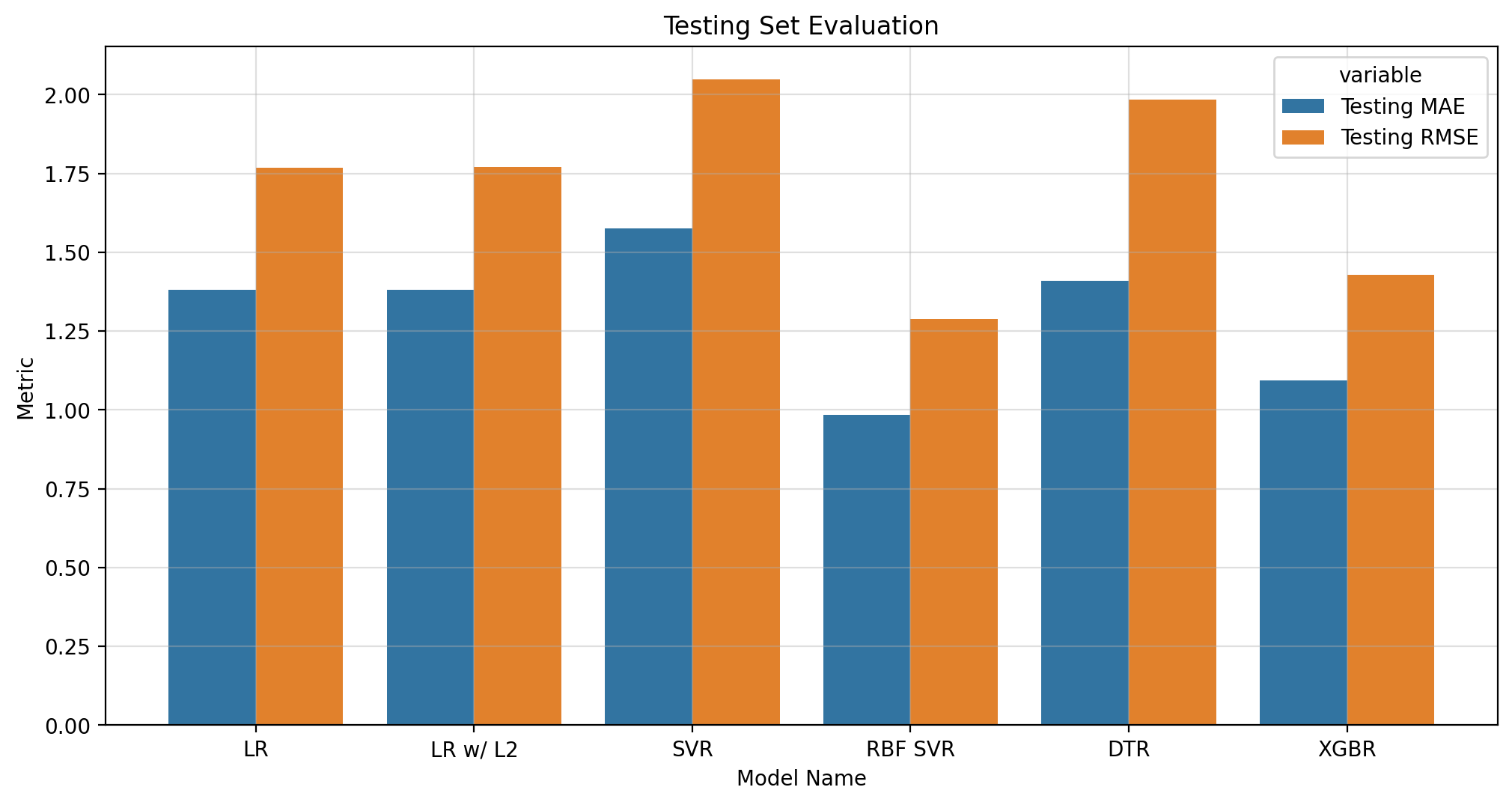}
    \captionof{figure}{Classical Modelling Performance}
    \label{fig:classical}
\end{center}

\subsection{Deep Learning}
\label{sec:deep}
In this section we apply deep learning algorithms to improve performance on the boulder grade prediction task. We evaluate the 2016 Moonboard Dataset and show increased performance with specialized architectures designed to incorporate spatial understanding of the distance between holds, then evaluate their ability to generalize between Moonboard configurations.

We evaluate a suite of 4 deep learning architectures: a 4-layer fully connected network (DNN), a 4-layer convolutional neural network (2DCNN), a single layer long-short term memory layer (LSTM) and a significantly deeper 6-layer fully connected network followed by 4-layer convolutional network(DNN+2DCNN). Exact architectures of each network are included as defined in the Keras API in Abstract 1. All networks were trained with the popular Adam optimizer and used a Mean Squared Error loss function. Networks were trained for 100 epochs with Early Stopping once the network had not improved on the test set in 20 epochs, with the best weights restored before evaluation.

Training and evaluating on the 2016 Moonboard Dataset, we find that the 4 deep-learning architectures outperform the RBF SVM. As shown in Figure \ref{fig:deep}, the 2DCNN achieves best performance with 0.86 MAE and 1.12 RMSE. This architecture was designed to learn spatial relationships between holds in local areas through the use of convolutional filters. We also evaluate the accuracy and within 1 grade accuracy of this model at 42\% and 84\% respectively. While each of the 4 convolutional layers all used a 3x3 kernel, as the feature vector propogates through the network each convolutional layer aggregates information from increasingly large local areas. While the first layer only has information about a 3x3 region, the second combines information from a 5x5 region, then 7x7 and so forth.

\begin{center}
    \centering
    \includegraphics[width=\linewidth]{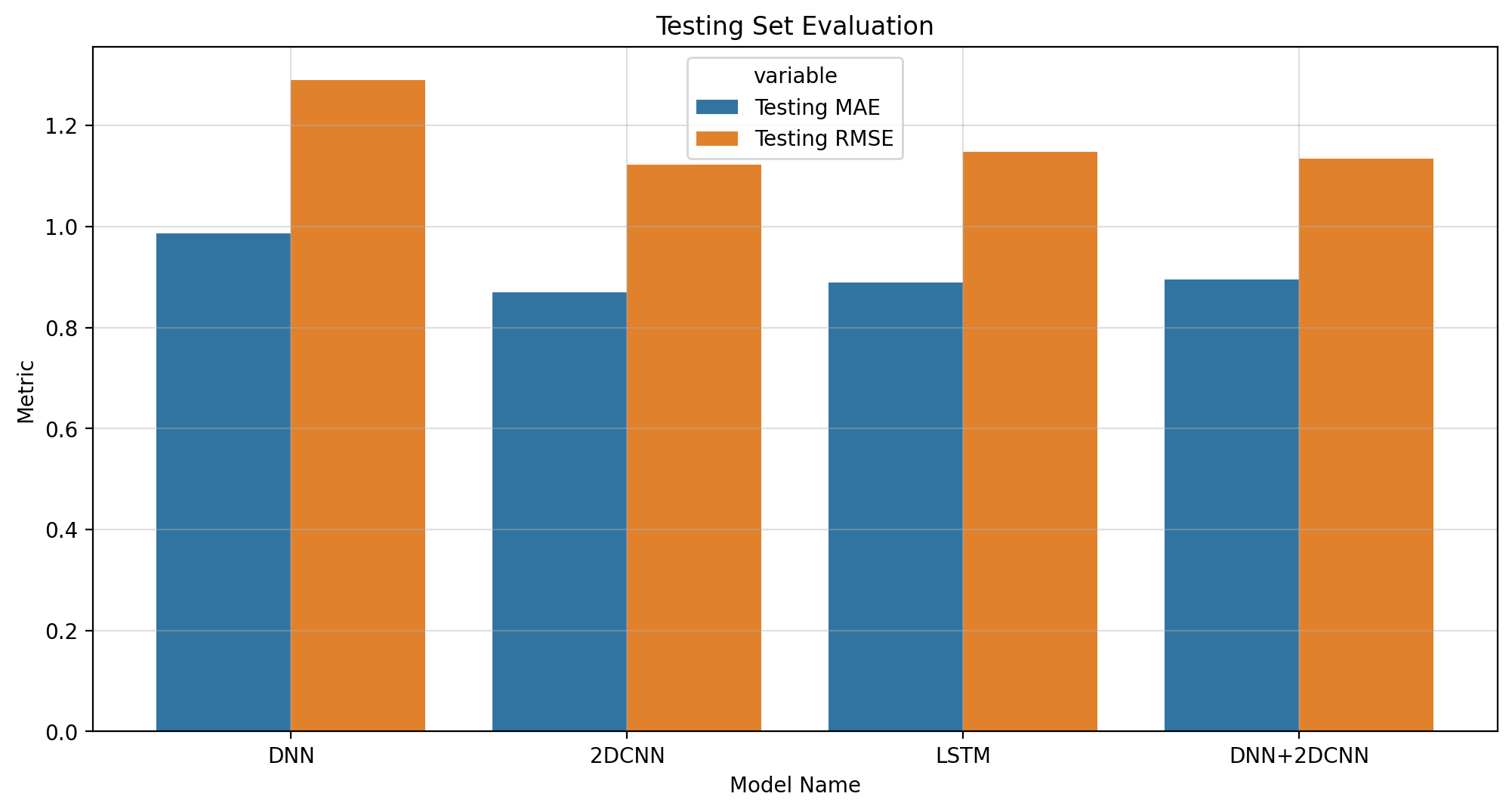}
    \captionof{figure}{Deep Learning Modelling Performance}
    \label{fig:deep}
\end{center}

Slightly lower performance was achieved by the LSTM network. This network was designed to process rows of the route as timesteps starting from the bottom row and moving upwards. For all architectures, we experimented with the addition of a class balancing weight to address the tendency for models to underestimate the grade, as shown in the residual plot in Figure \ref{fig:cnn residual}. This underestimation is likely due to the high class imbalance towards lower grades in the 2016 Dataset, as shown in Figure \ref{fig:distributions}. The class balancing weight multiplied loss during training by the inverse of that classes proportion in the training set. While this improved balance in the predictions, overall MAE and RMSE scores were similar.

\begin{center}
    \centering
    \includegraphics[width=\linewidth]{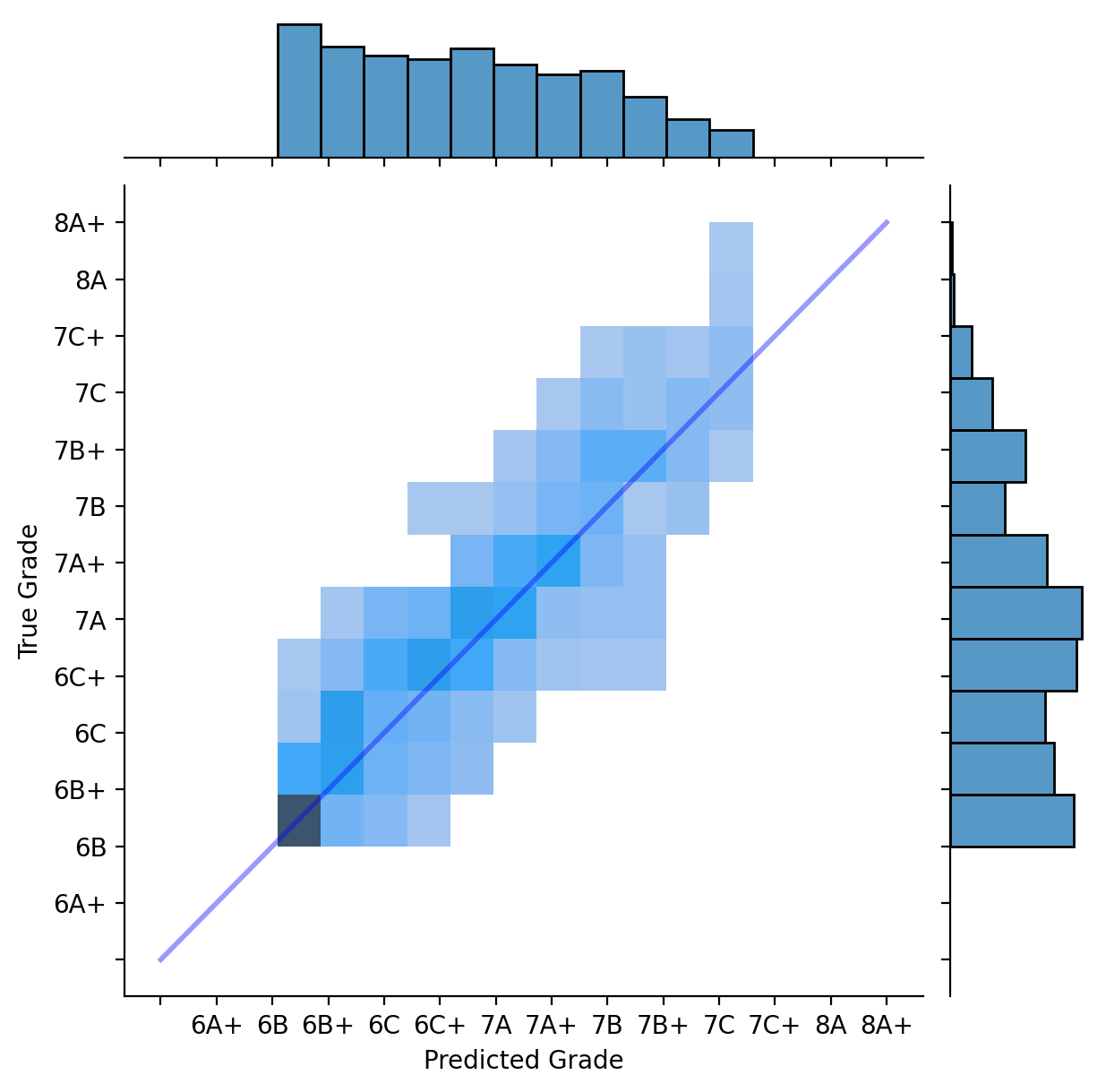}
    \captionof{figure}{CNN Architecture Residual Plot on Moonboard 2016 Benchmarks}
    \label{fig:cnn residual}
\end{center}

To address the relatively large number of routes with only few ascents, we experimented with the addition of individual sample weights equal to the log number of ascents of each route. Our assumption was that routes with many ascents would have more accurate grades and should therefore be weighted more heavily during training. Again, overall MAE and RMSE measurements were unaffected.

\begin{figure*}[t!]
    \centering
    \begin{subfigure}[t]{0.4\linewidth}
        \centering
        \includegraphics[height = 8cm]{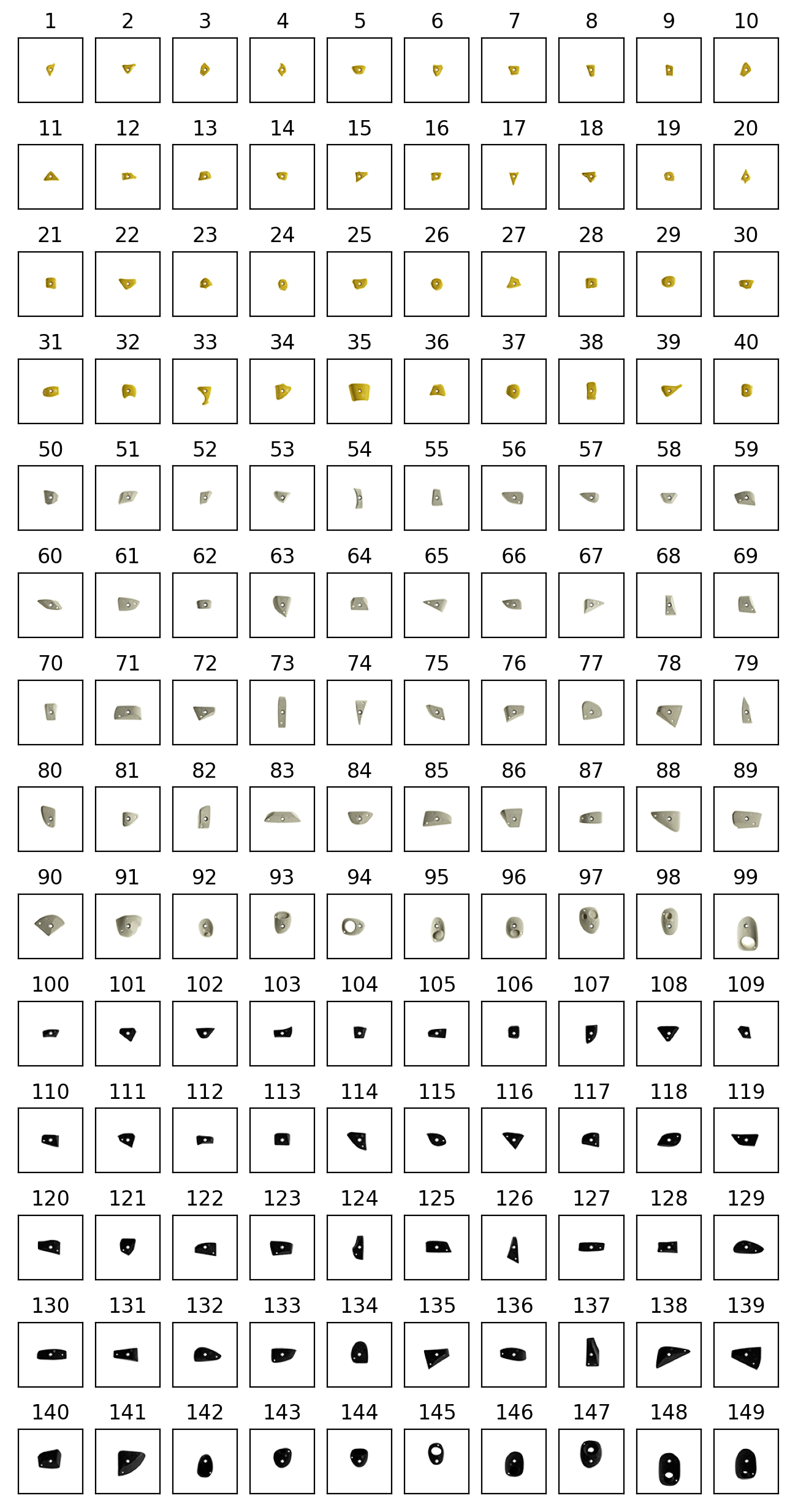}
        \caption{2016 Moonboard hold images before relocation and reorientation.}
    \end{subfigure}%
    ~ 
    \begin{subfigure}[t]{0.4\linewidth}
        \centering
        \includegraphics[height = 8cm]{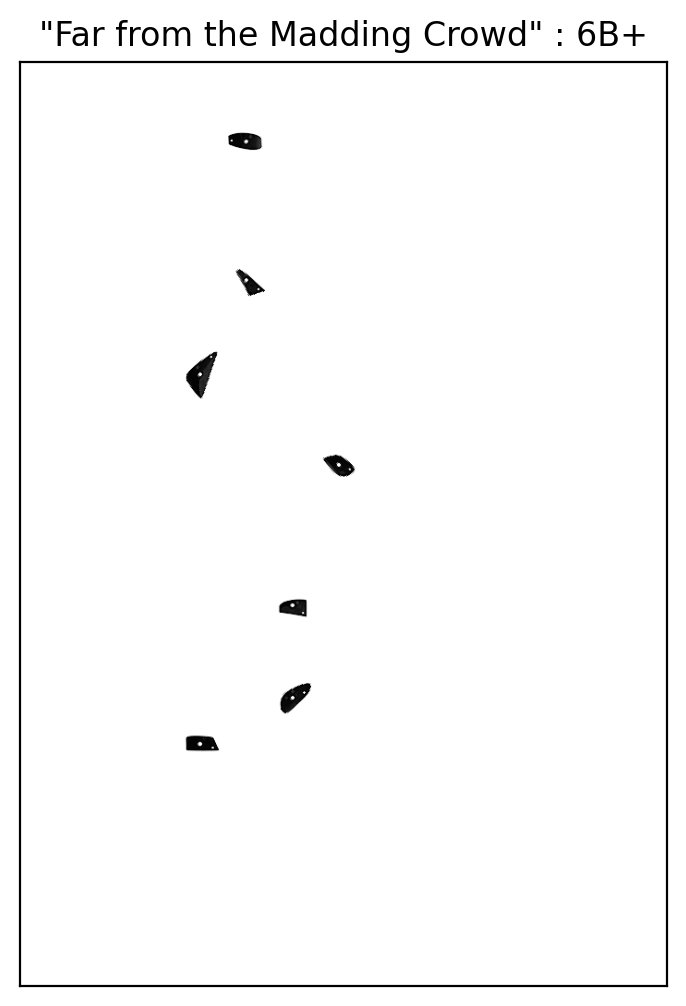}
        \caption{Sample image of a route generated from the 2016 Moonboard dataset. Each hold has been reoriented only the included holds in the route are displayed.}
    \end{subfigure}
    \captionof{figure}{Image Generation from 2016 Moonboard Dataset}
    \label{fig:image generation}
\end{figure*}

\subsection{Generalization of Modelling}
Next, we explore how deep learning modelling generalized between different editions of the Moonboard. The same suite of 4 models from Section \ref{sec:deep} were trained on one dataset, and evaluated against each of the three testing sets, as shown in Figure \ref{fig:generalization}.

Figure \ref{fig:generalization} shows that the CNN and LSTM architectures generalized best, perhaps due to their structure being designed to exploit spatial attributes of our feature set. Performance against unseen configurations was significantly lower than the configuration used during training with best performance achieved by the LSTM architecture. Worst generalization was achieved when training on the 2016 dataset. Possible explanations for this could be the relatively few number of holds on this configurations when compared to the 2017 and 2019 boards and the higher minimum grade of 6B+ compared to the 6A+ minimum on the later editions.

\begin{center}
    \centering
    \includegraphics[width=\linewidth]{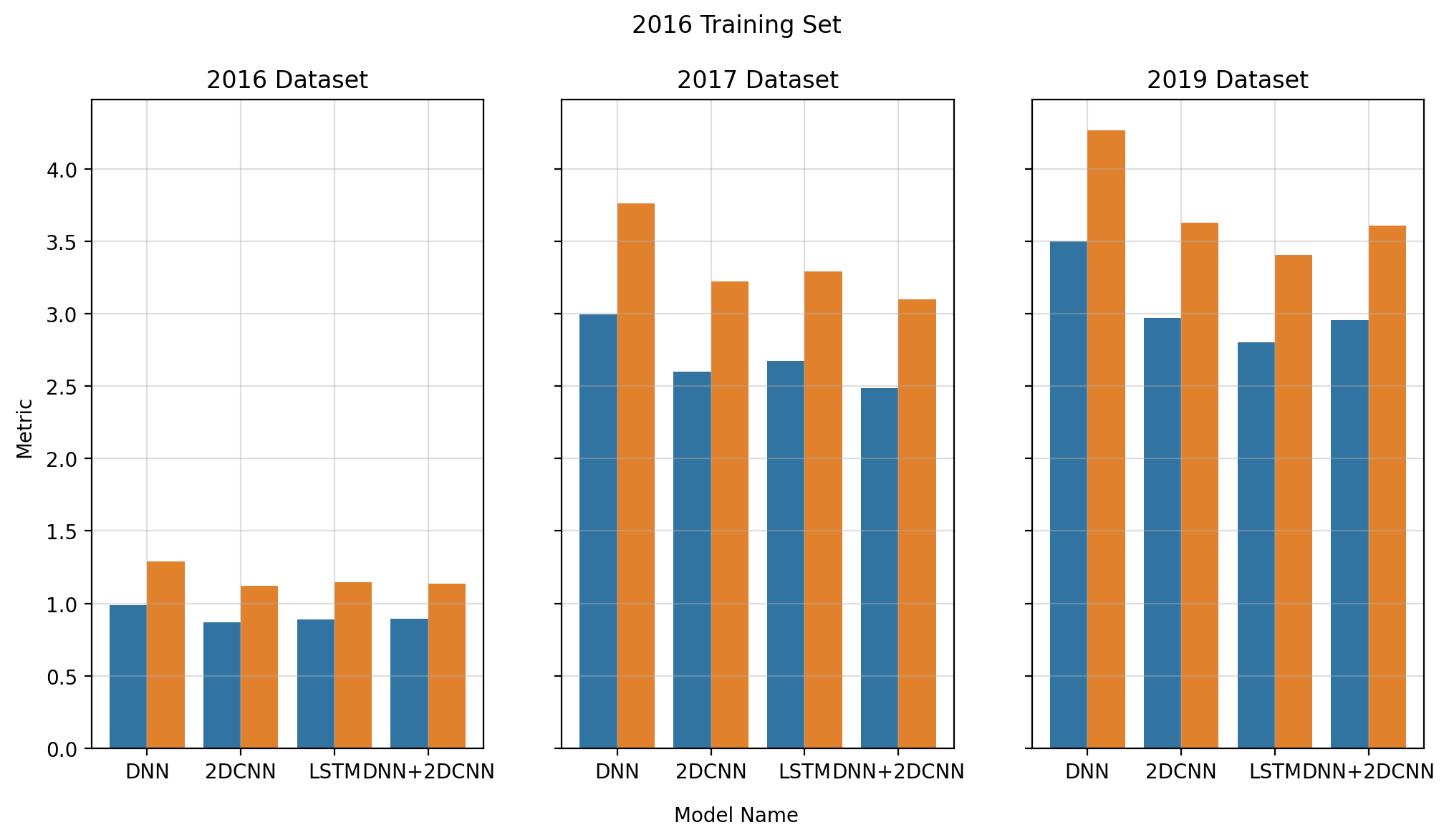}
    \includegraphics[width=\linewidth]{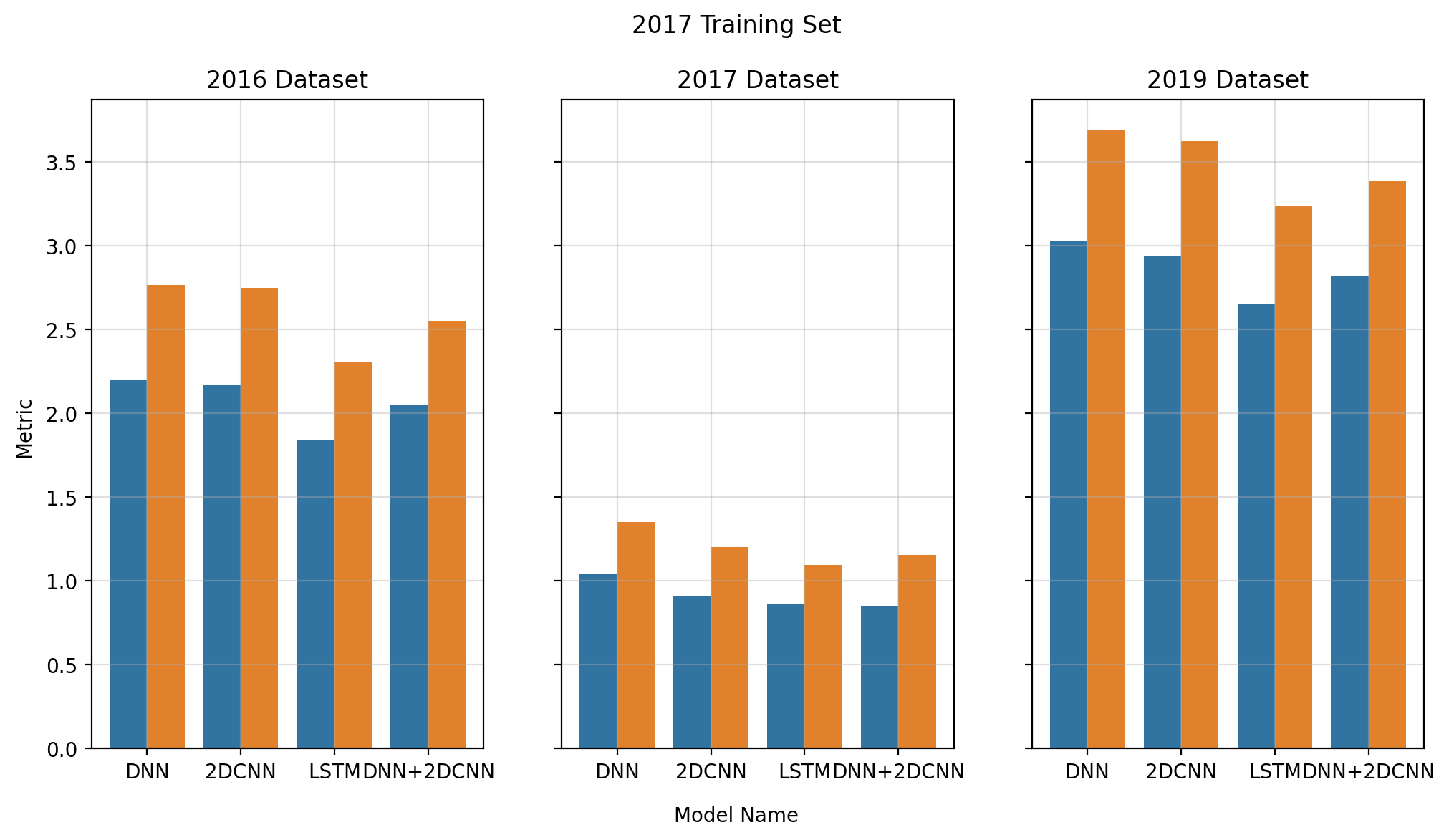}
    \includegraphics[width=\linewidth]{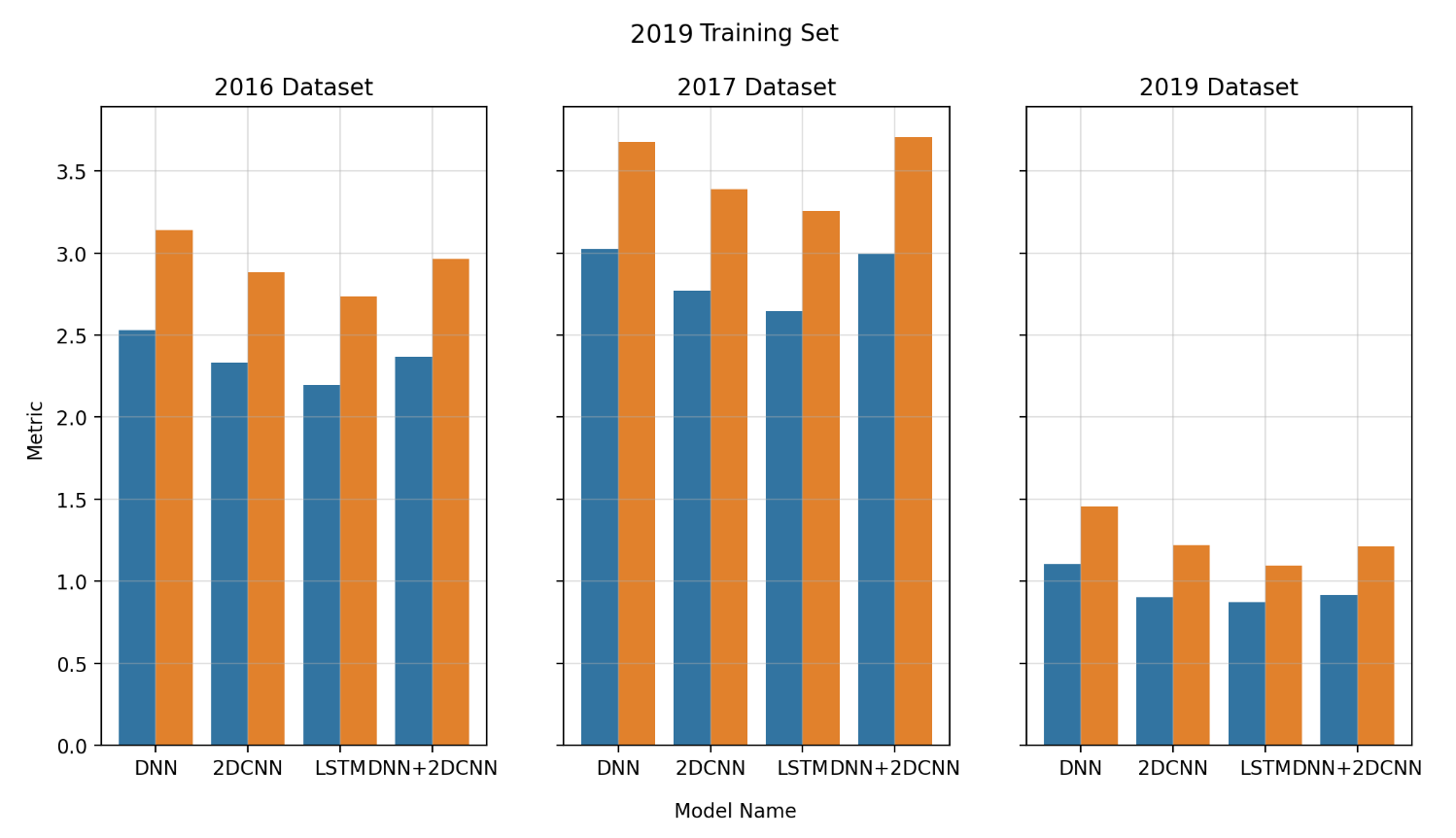}
    \captionof{figure}{Generalization Performance of Deep Learning Architectures against different Moonboard Edition Benchmarks}
    \label{fig:generalization}
\end{center}

A concern when evaluating generalization ability of a model trained on a single Moonboard edition is the overfitting of that model to specific hold positions. By varying the holds in each position, a model may be forced to learn spatial relationships rather than overfitting to individual positions. In Figure \ref{fig:generalization2}, we train the same suite of 4 deep learning models with two editions of the Moonboard and evaluate generalization performance to the third. We find that generalization performance does improve when compared to the single training set, however performance is significantly lower than the editions where the training set was directly included.

\begin{center}
    \centering
    \includegraphics[width=\linewidth]{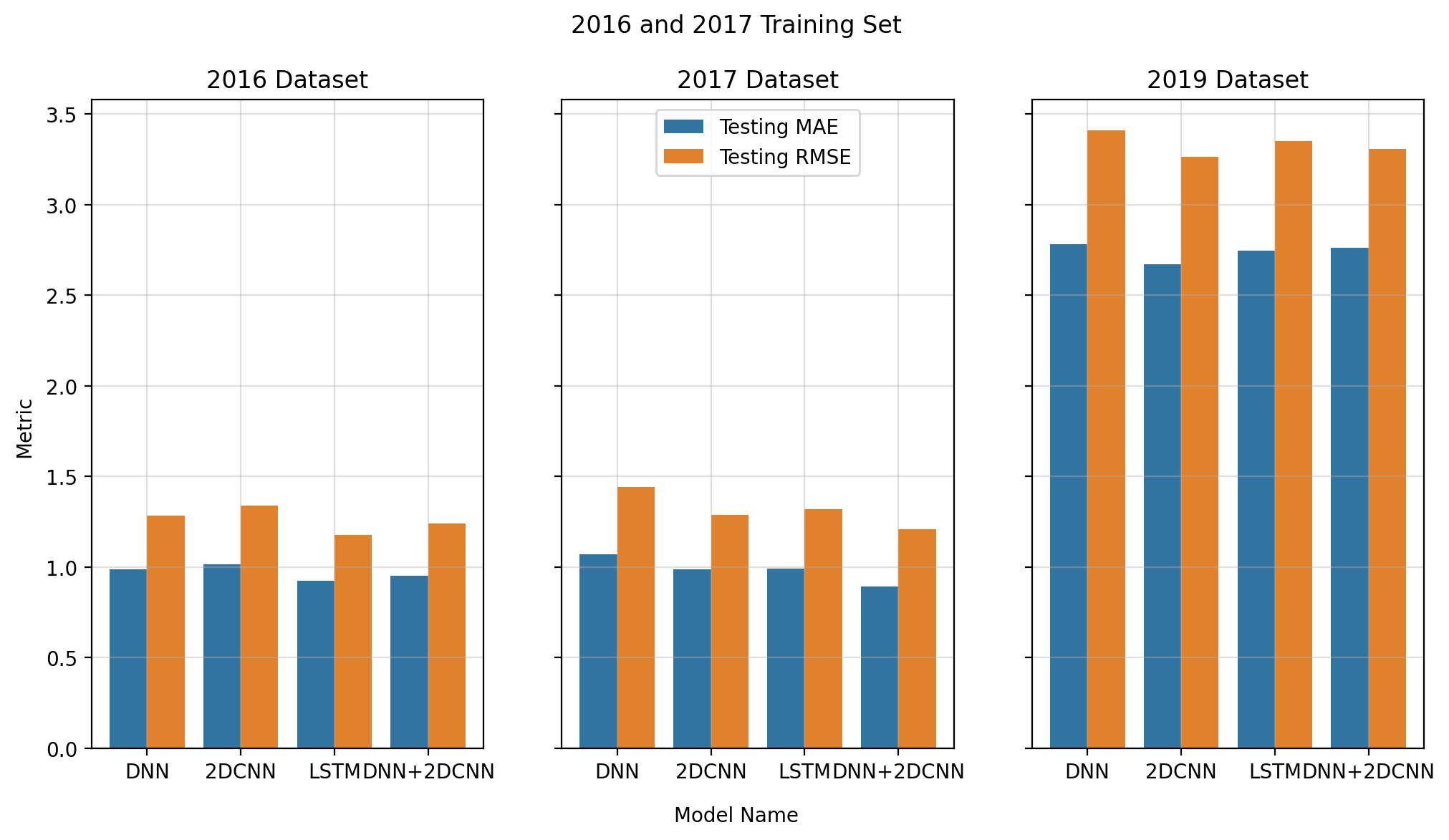}
    \includegraphics[width=\linewidth]{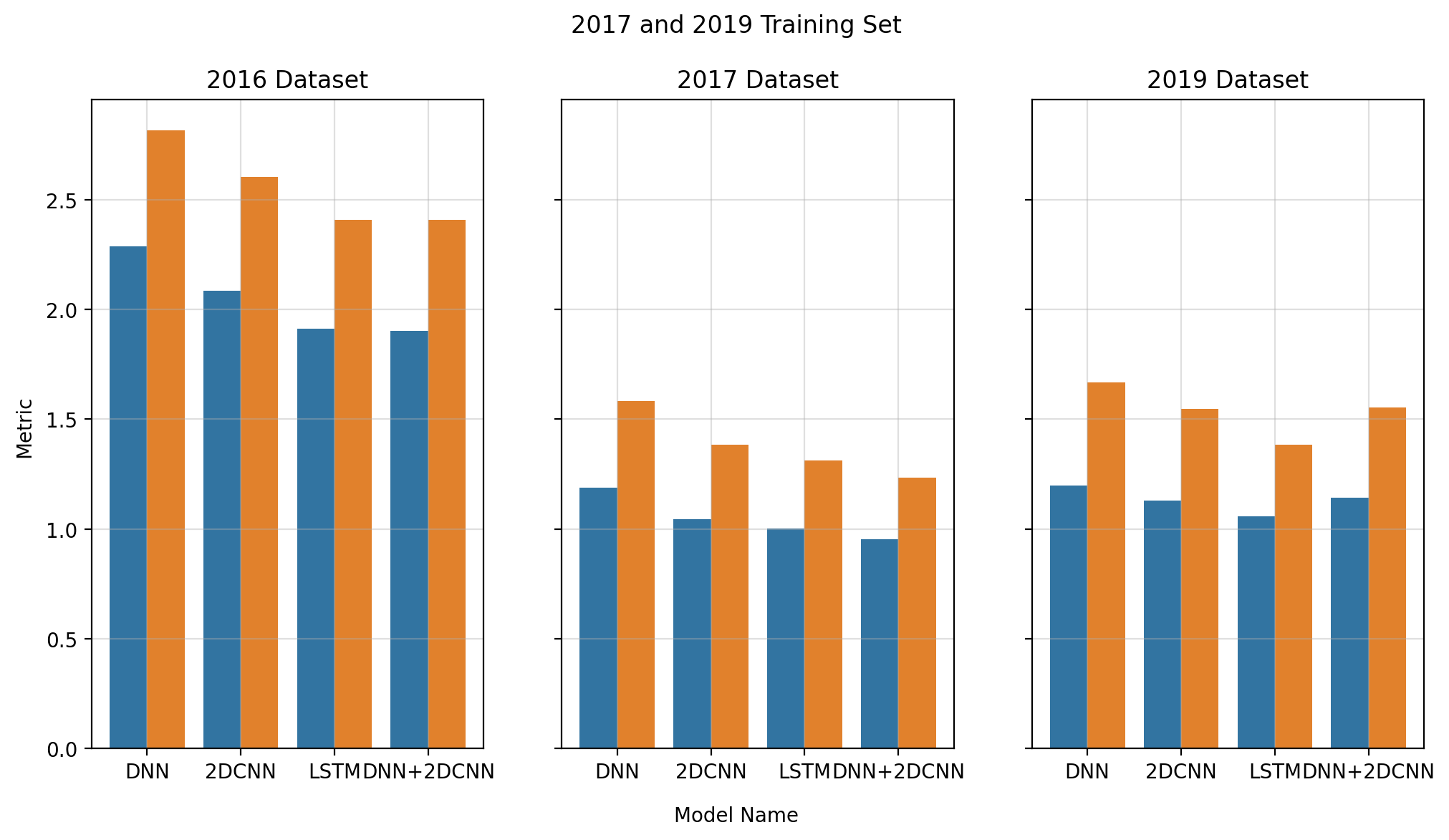}
    \includegraphics[width=\linewidth]{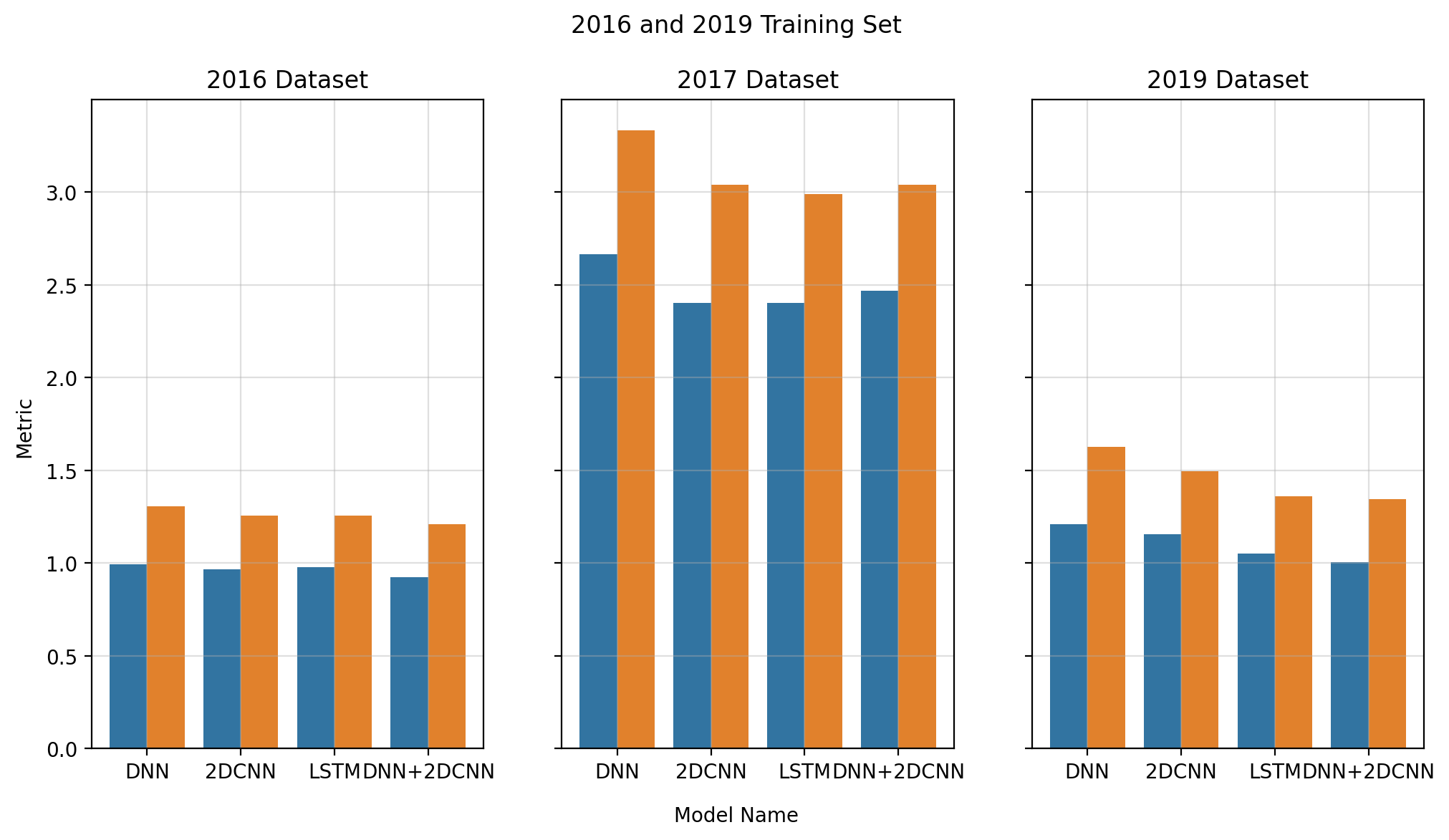}
    \captionof{figure}{Generalization Performance of Deep Learning Architectures against different Moonboard Edition Benchmarks}
    \label{fig:generalization2}
\end{center}

Figure \ref{fig:generalization2} also shows that testing set performance on editions that were included in training was reduced. This was most significant on the 2017 and 2019 training set, where best MAE and RMSE were approximately 0.3 higher than the single training set scores. With the one-hot encoded feature set, there is no way for the model to distinguish which edition a sample came from which could explain this reduced performance. Future work could explore generalization using the same hold set reorganised on a constant size board, which could lead to increased performance.

\section{Grade Prediction from Images}
A significant limitation of our method is the assumption of consistent holds equally spaced on a flat plane. Modern bouldering routes use a wide range of holds with irregular spacing and rotation, and are often set on complex 3D surfaces. Such complex orientation is difficult to capture in a consistent set of fixed-length features necessary for our modelling. This complexity motivated the use of images of routes to predict the grade. Such a method could generalize to images of any route and produce a predicted grade.

Collecting individual hold images, locations and orientations from the Moonboard website, a script was developed to generate images of routes using their included holds, as shown in Figure \ref{fig:image generation}.

Using the 2016 Moonboard dataset, an image for each route was generated and supervised with the route's grade. This image feature set was used to train both ResNet50 and MaxViT backbones. With both randomly initialized and default weights these architectures were unable to achieve performance on par with the original one-hot encoded feature space. Best performance was achieved by ResNet50 with no weights with 1.84 MAE and 2.30 RMSE.

We experimented with 224x224, 256x256 and 512x512 image sizes with no performance increase, as well as varying the number of channels in each image. Models were trained with single, three and four channel (monochrome, RGB and RGBA) images again to no improvement. This training was motivated by the hypothesis that black colored holds may be confused with the background black color signal for single and three channel images.

The ResNet50 model was also trained with categorical crossentropy loss rather than mean squared error to experiment with whether different loss functions lead to increased performance. Again, performance was poor with 2.25 MAE and 3.04 RMSE. Due to low performance on the same board edition during testing, we did not explore the generalization performance of this method.

\section{Conclusion}
Boulder grades are decided by climber's subjective opinions of the routes difficulty, and as such can be biased and inaccurate. In this work, we show that classical and deep-learning modelling can achieve highly accurate grade predictions on a standardised training board using the Moonboard dataset. Highest performance was achieved by the 2DCNN architecture's performance with a 0.86 MAE and 1.12 RMSE. By training on two editions of the board and evaluating generalization performance on a third, we demonstrate the LSTM achieves best performance with 2.35 MAE and 2.93 RMSE. While this demonstrates some generalization capability, this is far below estimated human level performance \cite{Duh2020RecurrentNN}\cite{stapel2023}. Finally, we experiment with vision based boulder grade prediction by generating images of routes from the 2016 Moonboard. While such a method could generalize to any board and hold set, future work is required to achieve human level performance from the current best losses of 1.84 MAE and 2.30 RMSE.

\section{Code Availability}
We make our code to generate images from Moonboard routes available for use here: \url{https://github.com/a1773620/Moonboard-Grade-Prediction}. We also publish our modelling techniques for classical, deep and computer vision techniques.

\newpage

\printbibliography

\section{Appendix}
\subsection{Deep Learning Architectures}
\begin{center}
    \centering
    \includegraphics[width=0.8\linewidth]{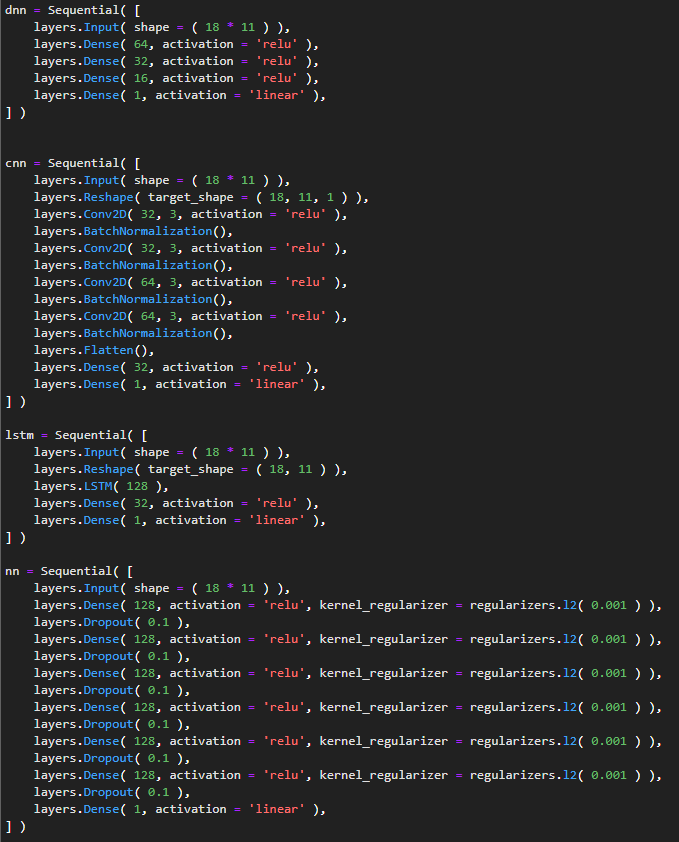}
    \captionof{figure}{Deep Learning architectures defined in Keras API}
    \label{fig:architectures}
\end{center}

\end{document}